# Kernel Cross-Correlator

**Chen Wang[1], Le Zhang[2], Lihua Xie[1], and Junsong Yuan[1]**
[1]School of Electrical and Electronic Engineering, Nanyang Technological University, Singapore
[2]Advanced Digital Sciences Center, Singapore
wang.chen@zoho.com, zhang.le@adsc.com.sg, {elhxie, jsyuan}@ntu.edu.sg

## Abstract

Cross-correlator plays a significant role in many visual perception tasks, such as object detection and tracking. Beyond the linear cross-correlator, this paper proposes a kernel cross-correlator (KCC) that breaks traditional limitations. First, by introducing the kernel trick, the KCC extends the linear cross-correlation to non-linear space, which is more robust to signal noises and distortions. Second, the connection to the existing works shows that KCC provides a unified solution for correlation filters. Third, KCC is applicable to any kernel function and is not limited to circulant structure on training data, thus it is able to predict affine transformations with customized properties. Last, by leveraging the fast Fourier transform (FFT), KCC eliminates direct calculation of kernel vectors, thus achieves better performance yet still with a reasonable computational cost. Comprehensive experiments on visual tracking and human activity recognition using wearable devices demonstrate its robustness, flexibility, and efficiency. The source codes of both experiments are released at https://github.com/wang-chen/KCC.

## Introduction

Cross-correlator (CC) or correlation filter (CF) usually refers to the function or device to compute cross-correlation for signals. It has been widely used in many visual perception tasks such as object detection, recognition, and tracking. The recent extensions in visual tracking, which are usually named as correlation filters, e.g., minimum output sum of squared error (MOSSE) (Bolme et al. 2010), rekindle the interest of real-time applications due to their robust translation prediction. One of the main advantages of CF is its efficient calculation with a complexity of $\mathcal{O}(n \log n)$ in Fourier domain versus $\mathcal{O}(n^3)$ in spatial domain, where $n$ is the length of a signal. The kernelized correlation filter (KCF) extends MOSSE to kernel case and obtains impressive performance on robust object tracking (Henriques, Caseiro, and Martins 2015). Discriminative scale space tracking (DSST) (Danelljan et al. 2017) learns separate CFs on different scales, enabling estimation of both the translation and scale variants.

In spite of the theoretical improvements, the existing CFs are still limited in some aspects. First, only the translational changes of the input can be estimated by viewing the corresponding movement of the maixmum output. The estimation for other transformations are usually achieved by combining multiple CFs with different input transforms, e.g. the scale filters in (Danelljan et al. 2017). Second, KCF extends CF to non-linear case but is only applicable for training data with circulant structure (each sample is circular translation of others) and non-weighted kernel functions, which limit its usage in other applications where those assumptions turn out to be invalid (Henriques, Caseiro, and Martins 2015). To address these problems, a **kernel cross-correlator** (KCC) is proposed. In summary, the main contributions are:

- By introducing the kernel trick, the KCC extends the cross-correlation to non-linear space, which is more robust to signal noises and distortions;
- Compared to KCF, KCC can be applied to any data structure using any kernel functions, hence its properties are customizable and can be used to predict any affine transformations, especially the rotation, scale, and translation;
- KCC generalizes the existing CFs and provides unified solutions, which improves the performance of visual tracking while still keeps efficiency;
- KCC provides a similarity measurement for signals with transformation. The extensive experiments on human activity recognition using wearable devices demonstrate its robustness and flexibility.

Without loss of generality, the name ‘cross-correlator’ is preferred in this paper rather than ‘correlation filter’, since the former is designed not only for filtering purpose. Considering that the CFs have shown their great capabilities in numerous applications including template matching, visual tracking, face verification, etc., we believe that KCC will also play an important role in those fields.

## Related Work

The wealth of research in this area is such that we cannot give an exhaustive review. Instead, only several milestones on CFs and visual tracking are presented.

### Correlation Filter

One of the earliest correlation filters is the synthetic discriminant filter (SDF) that describes a signal as a linear combination of orthonormal basis function expansion, so that a single averaged spatial filter can be obtained from a weighted linear combination of these functions (Hester and Casasent 1980). The unified synthetic discriminant filters (USDF) provide a general basis function and hyperspace description for SDFs and show the generality of the correlation matrix observation space (Casasent 1984). Later, it is proved that SDF is only suitable for processing white noises and a minimum-variance synthetic discriminant function (MVSDF) is designed for colored noises (Kumar 1986). To obtain sharper correlation peaks, the minimum average correlation energy (MACE) filter proposes to minimize the energy of the correlation output and this facilitates the translational target detection (Mahalanobis, Kumar, and Casasent 1987). To ensure the sharpness of correlation peak against noise robustness and horner efficiency, the optimal trade-off filter (OTOF) is designed (Refregier 1991).

Afterwards, researchers' interests have been focused on distortion performance. For instance, the unconstrained correlation filter (UCF) removes the usual hard constraints on the output of SDF (Mahalanobis et al. 1994). This exhibits superior distortion tolerance while still keeps the attractive features of MACE and MVSDF. The distance-classifier correlation filter (DCCF) enables the mutli-class recognition by introducing a distance-classifier scheme (Mahalanobis, Vijaya Kumar, and Sims 1996). By specifying the desired response at every location, the average synthetic exact filter (ASEF) generalizes across entire training set by averaging multiple exact filters (Bolme, Draper, and Beveridge 2009). To overcome the overfitting problem of ASEF, the minimum output sum of squared error (MOSSE) filter adds a regularization term and introduces it into visual tracking (Bolme et al. 2010). Its superior speed and robustness ignite the boom and development of CF based tracking.

Kernelized correlation fiter (KCF) extends the linear filters to non-linear space by introducing the kernel trick into ridge regression (Henriques, Caseiro, and Martins 2015). This enables learning with element-wise operation instead of costly matrix inversion, providing much more robustness while still with reasonable learning speed. Multi-kernel correlation filter (MKCF) (Tang and Feng 2015) takes advantage of the invariance-discriminative power spectrums of various features to further improve the performance. To alleviate the boundary effect of CFs, zero aliasing correlation filter (ZACF) (Fernandez et al. 2015) introduces the zero-aliasing constraints and provides both closed-form and iterative proximal solutions by ensuring that the optimization criterion for a given CF corresponds to a linear correlation rather than a circular correlation. However, it requires heavy computation and is not suitable for real-time applications. Discriminative scale space tracking (DSST) (Danelljan et al. 2017) is proposed to learn separate MOSSE on different scales, enabling estimation of both translation and scale variants, which is at the cost of repeated calculations of MOSSE.

### Discriminative Tracking

Discriminative trackers have received wide attention in recent years and have established its superiority over generative ones (Avidan 2007; Babenko, Yang, and Belongie 2011). Along with the established discriminative trackers which use advanced off-the-shelf machine learning algorithms such as multiple-instance boosting (Babenko, Yang, and Belongie 2011), kernelized structured SVM (Hare, Saffari, and Torr 2011), graph learning (Li et al. 2017), self-supervised learning (Wu and Huang 2000), dictionary based trackers (Mei et al. 2013), several ensemble methods such as boosting (Grabner, Grabner, and Bischof 2006) and random forest were proposed (Zhang et al. 2017).

The deep learning based discriminative trackers have also been extensively studied. In (Wang and Yeung 2013), the authors make the first attempt to transfer rich feature hierarchies from a large auxiliary dataset based on an auto-encoder. The work in (Wang et al. 2015a) addresses the problem of online training and finetuning deep convolutional neural network for visual tracking. (Wang et al. 2016) formulates convolutional neural network as ensemble learning and shows its effectiveness in visual tracking. Tracking by convolutional neural network without training can be found in (Zhang et al. 2016; Zhang and Suganthan 2016).

As aforementioned, the recent trend towards deep learning based discriminative trackers mitigates tracking error to an extent, but the tracking speed may still be low. This challenge opens space for learning techniques that aim to achieve higher accuracy and yield faster training. To this end, the correlation filters based tracking MOSSE (Bolme et al. 2010) has been proposed with extremely high computational efficiency. Since then, numerous extensions for correlation filter based trackers, such as KCF, DSST, and Staple (Bertinetto et al. 2016), have been proposed which outperform other trackers in this family. In (Liu, Wang, and Yang 2015), part based KCF is proposed to address the problem of occlusion. The reliable patch tracker (RPT) presents a tracking reliability metric to measure how reliably a patch can be tracked, where a probability model is proposed to estimate the distribution of reliable patches under a sequential Monte Carlo framework (Li, Zhu, and Hoi 2015). The multiple kernel correlation filter (MKCF) tracker takes advantage of the invariance-discriminative power spectra of various features to further improve the performance (Tang and Feng 2015). An online random fern classifier is employed as a re-detection component for the long-term correlation tracking (LCT) (Ma et al. 2015b). A biology-inspired framework where short-term processing and long-term processing are cooperated with each other in correlation filter based trackers (Hong et al. 2015). In (Ma et al. 2015a), the authors hierarchically infer the maximum response of ensemble of correlation filters trained on the features extracted from different layers of convolutional neural networks (ConvNet). Fusion with CF based tracker with different features can be found in (Zhang and Suganthan 2017).

## Formulation

Unlike correlation which is associated with dependence in statistics, cross-correlation is defined against convolution in signal processing. In this section, a new formulation for non-linear cross-correlator based on kernel trick is presented. For the sake of simplicity, signals will be denoted as a column vector $\mathbf{z}, \mathbf{x} \in \mathbb{R}^n$. The derived results can be easily extended to matrix case. The convolution theorem states that cross-correlation becomes element-wise conjugate multiplication in Fourier domain. Denote the fast Fourier transform (FFT) $\mathcal{F}: \mathbb{C}^n \mapsto \mathbb{C}^d$ on a vector as $\hat{\cdot}$, so that the cross-correlation of two vectors $\mathbf{g} = \mathbf{x} * \mathbf{h}$ is equivalent to $\hat{\mathbf{g}} = \hat{\mathbf{x}} \odot \hat{\mathbf{h}}^*$, where the operator $\odot$ and superscript $^*$ denote the element-wise multiplication and complex conjugate, respectively. Then the output $\hat{\mathbf{g}}$ can be transformed back into the spatial domain using the inverse FFT. Therefore, the bottleneck of cross-correlation is to compute the forward and backward FFTs, and the complexity of entire process has an upper bound $\mathcal{O}(n \log n)$. The cross-correlator is usually a learning problem, whose objective is to encode the pattern of training samples into a correlator $\mathbf{h}$, so that the pattern of test samples can be reflected by the correlation output $\mathbf{g}$.

The kernel cross-correlator is defined basesd on the kernel trick. For a non-linear feature mapping function $\varphi : \mathbb{R}^n \mapsto \mathbb{R}^d$, $d \gg n$, the kernel trick is to find the inner product of feature mapping without calculating the high dimension features explicitly. Denote the kernel function as $\kappa : \mathbb{R}^n \times \mathbb{R}^n \mapsto \mathbb{R}$, such that $\kappa(\mathbf{x}_i, \mathbf{x}_j) = \varphi(\mathbf{x}_i)^T \varphi(\mathbf{x}_j)$. Given a signal $\mathbf{x} \in \mathbb{R}^n$ and desired output $\mathbf{g} \in \mathbb{R}^m$, the kernel cross-corrrelator is defined as:

$$\boxed{\hat{\mathbf{g}} = \hat{\boldsymbol{\kappa}}_{\mathbf{z}}(\mathbf{x}) \odot \hat{\mathbf{h}}^*}, \tag{1}$$

where $\boldsymbol{\kappa}_{\mathbf{z}}(\mathbf{x})$ is the kernel vector. For single kernel case, $\boldsymbol{\kappa}_{\mathbf{z}}(\mathbf{x}) = [\kappa(\mathbf{x}, \mathbf{z}_1), \cdots, \kappa(\mathbf{x}, \mathbf{z}_m)]^T$, where $\mathbf{z}_i \in \mathcal{T}(\mathbf{z}) \in \mathbb{R}^n$ is generated from the training sample $\mathbf{z}$. The transform function $\mathcal{T}(\cdot)$ is predefined to imply the training objective. For multi-kernel case, each element of the kernel vector $\boldsymbol{\kappa}_{\mathbf{z}}(\mathbf{x})$ is a linear combination of a set of kernels $\kappa_i$, where $i = 1, 2, \cdots l$. The $k_{th}$ element of the kernel vector is defined as $\mathbf{d}^T \mathbf{k}_k$, where $\mathbf{d} \in \mathbb{R}^l$ is the weights vector subject to $\|\mathbf{d}\|_1 = 1$, and the vector stack $\mathbf{k}_k = [\kappa_1(\mathbf{x}, \mathbf{z}_k), \cdots, \kappa_l(\mathbf{x}, \mathbf{z}_k)]^T$.

Note that the length of $\mathbf{h}$ and the target $\mathbf{g}$ are related to the number of sample-based vectors $\mathbf{z}_i$, i.e. $\mathbf{h}, \mathbf{g} \in \mathbb{R}^m$ and normally $n \neq m$. The desired output $\mathbf{g}$ can take any shapes, e.g. a Gaussian function with the peak centered on the training signal. In the training stage, $\mathbf{z}_i$ can take any affine transformation of $\mathbf{z}$. The corresponding patterns will be encoded into $\mathbf{h}$, so that the pattern of a test sample $\mathbf{x}$ can be predicted through the correlation output $\mathbf{g}(\mathbf{x})$. For example, if the desired output $\mathbf{g}$ is a binary vector and the single peak locates on the first element, the location of the maximum response in the correlation output $\mathbf{g}(\mathbf{x})$ indicates the test sample's pattern associated with that location in the kernel vector $\boldsymbol{\kappa}_{\mathbf{z}}(\mathbf{x})$.

Denote $\mathbf{z}^i$ as the $i_{th}$ training sample. Note that $\mathbf{z}^i$ is different from the sample-based vector $\mathbf{z}_i$ generated from the training sample $\mathbf{z}$. To be efficient, training is conducted in the Fourier domain to take advantage of simple element-wise operation. The training objective is to find $\mathbf{h}$ that maps $\mathbf{z}^i$ to the desired output $\mathbf{g}^i$ and minimizes the sum of squared error (SSE) between the output and desired output:

$$\min_{\hat{\mathbf{h}}^*} \sum_{i=1}^{s} \|\hat{\boldsymbol{\kappa}}_{\mathbf{z}^i}(\mathbf{z}^i) \odot \hat{\mathbf{h}}^* - \hat{\mathbf{g}}^i\|^2 + \lambda \|\hat{\mathbf{h}}^*\|^2, \tag{2}$$

where $s$ is the number of training samples. For single kernel case, $\hat{\boldsymbol{\kappa}}_{\mathbf{z}^i}(\mathbf{z}^i) = [\kappa(\mathbf{z}^i, \mathbf{z}_1^i), \cdots, \kappa(\mathbf{z}^i, \mathbf{z}_m^i)]^T$ and $\mathbf{z}_j^i$ is generated from $\mathbf{z}^i$. The second term in (2) is a regularization to prevent overfitting. Solving the optimization problem (2) requires setting the first derivative of $\mathbf{h}^*$ to zero, i.e.,

$$\frac{\partial}{\partial \hat{\mathbf{h}}^*} \left( \sum_{i=1}^{s} \|\hat{\boldsymbol{\kappa}}_{\mathbf{z}^i}(\mathbf{z}^i) \odot \hat{\mathbf{h}}^* - \hat{\mathbf{g}}^i\|^2 + \lambda \|\hat{\mathbf{h}}^*\|^2 \right) = 0. \tag{3}$$

Since all the operations in (3) are performed in element-wise, the elements of $\hat{\mathbf{h}}^*$ can be solved independently and a closed-form solution is obtained:

$$\boxed{\hat{\mathbf{h}}^* = \frac{\sum_{i=1}^{s} \hat{\mathbf{g}}^i \odot \hat{\boldsymbol{\kappa}}^*_{\mathbf{z}^i}(\mathbf{z}^i)}{\sum_{i=1}^{s} \hat{\boldsymbol{\kappa}}_{\mathbf{z}^i}(\mathbf{z}^i) \odot \hat{\boldsymbol{\kappa}}^*_{\mathbf{z}^i}(\mathbf{z}^i) + \lambda}}, \tag{4}$$

where the operator $\div$ denotes the element-wise division. Until now, the KCC is obtained. Since the kernel functions and the sample-based vectors $\mathbf{z}_i$ are not specified in the training process, any kernel and training data can be applied. This generalizes the existing CFs fundamentally, e.g. MOSSE, KCF, MKCF, and DSST, which will be presented later. More specifically, the properties of KCC can be customized, if the transform function $\mathcal{T}(\cdot)$ and the kernel functions $\kappa(\cdot, \cdot)$ are specified, which will be presented in the next section.

## Special Cases of KCC

The properties of KCC are customizable when the transform function $\mathcal{T}(\cdot)$ is specified for the kernel vector $\boldsymbol{\kappa}_{\mathbf{z}}(\mathbf{x})$. This section presents three examples using the translation transform $\mathcal{T}_T(\cdot)$, rotation transform $\mathcal{T}_R(\cdot)$, and scale transform $\mathcal{T}_S(\cdot)$, resulting in the special cases of KCC, i.e. kernel translation correlator (KTC), kernel rotation correlator (KRC), and kernel scale correlator (KSC), respectively. Without loss of generality, the single radial basis kernel (5) is used, the cases for other kernels are similar.

$$\kappa(\mathbf{x}, \mathbf{z}_i) = h\left(\|\mathbf{x} - \mathbf{z}_i\|^2\right). \tag{5}$$

### Kernel Translation Correlator

For a training sample $\mathbf{z} \in \mathbb{R}^n$ with length $n$, the number of all possible translational shifts $|\mathcal{T}_T(\mathbf{z})| = n$, where the operator $|\cdot|$ returns the number of elements in a set. In this sense, the length of $\mathbf{h}$ and $\boldsymbol{\kappa}_{\mathbf{z}}(\mathbf{x})$ equals the length of input signal, i.e. $m = n$. In spatial domain, the complexity for calculating a kernel (5) is $\mathcal{O}(n)$, hence the complexity for computing a kernel vector $\boldsymbol{\kappa}_{\mathbf{z}}(\mathbf{x})$ is $\mathcal{O}(n^2)$. Fortunately, it will be shown that the calculation can be done in Fourier domain with complexity $\mathcal{O}(n \log n)$. First, we have

$$h\left(\|\mathbf{x} - \mathbf{z}_i\|^2\right) = h(\|\mathbf{x}\|^2 + \|\mathbf{z}_i\|^2 - 2\mathbf{x}^T \mathbf{z}_i). \tag{6}$$

Since $\|\mathbf{x}\|^2$ and $\|\mathbf{z}_i\|^2$ are constants, then

$$\boldsymbol{\kappa}_{\mathbf{z}}(\mathbf{x}) = \left[h\left(\|\mathbf{x} - \mathbf{z}_1\|^2\right), \cdots, h\left(\|\mathbf{x} - \mathbf{z}_n\|^2\right)\right]^T \tag{7a}$$

$$= h(\|\mathbf{x}\|^2 + \|\mathbf{z}\|^2 - 2\left[\mathbf{x}^T \mathbf{z}_1, \cdots, \mathbf{x}^T \mathbf{z}_n\right]^T). \tag{7b}$$

From the cross-correlation theory, we know that $\mathbf{x} * \mathbf{z} = \left[\mathbf{x}^T \mathbf{z}_1, \cdots, \mathbf{x}^T \mathbf{z}_n\right]^T$. Substitute it into (7b), then

$$\boldsymbol{\kappa}_{\mathbf{z}}(\mathbf{x}) = h(\|\mathbf{x}\|^2 + \|\mathbf{z}\|^2 - 2 \cdot \mathbf{x} * \mathbf{z}) \tag{8a}$$

$$= h(\|\mathbf{x}\|^2 + \|\mathbf{z}\|^2 - 2 \cdot \mathcal{F}^{-1}(\hat{\mathbf{x}} \odot \hat{\mathbf{z}}^*)). \tag{8b}$$

Since the bottle-neck of (8b) is the forward and backward FFTs, the kernel vector can be calculated in complexity $\mathcal{O}(n \log n)$. For implementation purpose, the vector norm in (8b) is obtained in frequency domain using Parseval's theorem, so that the original signals don't need to be stored.

$$\boldsymbol{\kappa}_{\mathbf{z}}(\mathbf{x}) = h(\frac{1}{n}\|\hat{\mathbf{x}}\|^2 + \frac{1}{n}\|\hat{\mathbf{z}}\|^2 - 2 \cdot \mathcal{F}^{-1}(\hat{\mathbf{x}} \odot \hat{\mathbf{z}}^*)). \quad (9)$$

Therefore, based on (8b) or (9), KCC can be used to predict the translation transform of a signal. This results in the kernel translation correlator (KTC). Note that other correlation filters, e.g. KCF, are also able to predict translation, but with many limitations. It will be shown that KCF is a special case of KTC, since it is only applicable for non-weighted kernels and single training sample.

### Kernel Rotation Correlator

The kernel rotation correlator (KRC) takes the rotation transformation $\mathcal{T}_R(\mathbf{z})$ to generate the sample-based vectors $\mathbf{z}_i$. In applications, the rotation transforms are usually discretized with specific resolution $\epsilon$, hence the number of transforms $|\mathcal{T}_R(\mathbf{z})| = 2\pi/\epsilon = m$ is a constant and is independent of signal length $n$. Therefore, the kernel vector can be calculated in spatial domain with complexity $\mathcal{O}(n)$.

Because of the periodic structure in the rotation space, i.e. $\phi = \phi + 2\pi$, where $\phi$ is a rotation angle, the kernel vector $\boldsymbol{\kappa}_{\mathbf{z}}(\mathbf{x})$ is cyclic when it is extended to more than one circle. Hence, there is no boundary effect for the rotation correlator, making it more accurate than the others. In some applications, the rotation angle between two consecutive frames is small, hence it is not necessary to calculate the full kernel vector $\boldsymbol{\kappa}_{\mathbf{z}}(\mathbf{x})$ at all positions. Instead, we only need to compute them within a specified range in (10), while setting all the others to 0. This further reduces the computational requirements, while still keeps the periodic property.

$$\boldsymbol{\kappa}_{\mathbf{z}}(\mathbf{x}) = [0, \cdots, \kappa(\mathbf{x}, \mathbf{z}_i), \cdots, 0]^T, \quad (10)$$

where $\mathbf{z}_i \in \mathcal{T}'_R(\mathbf{z})$ only contains the possible rotations. To the best of our knowledge, this is the first non-linear rotation correlator, which may be useful for many applications, such as the rotation prediction in visual tracking.

### Kernel Scale Correlator

Similarly, the sample-based vectors $\mathbf{z}_i$ are obtained from the scale transforms, and only a small set need to be calculated. However, different from the rotation, the periodic property is not kept in the scale space. Therefore, the boundary effects may have a negative impact on the performance. It is possible to use the existing methods to eliminate this effect (Fernandez et al. 2015; Galoogahi, Sim, and Lucey 2015), but it is out of the scope of this paper. One of the simplest methods is to add zero padding to the kernel vector,

$$\boldsymbol{\kappa}_{\mathbf{z}}(\mathbf{x}) = [\kappa(\mathbf{x}, \mathbf{z}_1), \cdots, \kappa(\mathbf{x}, \mathbf{z}_m), 0, \cdots, 0]^T. \quad (11)$$

In this case, the length of the kernel vector will be doubled, i.e., $\boldsymbol{\kappa}_{\mathbf{z}}(\mathbf{x}), \mathbf{h} \in \mathbb{R}^{2m}$. Some correlation filters, e.g. DSST and LCT are also able to predict scale transforms, however, it is achieved by repeatedly calculating the translation correlators at different scales, and the best scale transform is obtained by the maximum translation response among all scales. Therefore, they are not scale correlators in the strict sense. More details about this will be presented in next section. To the best of our knowledge, KST is the first nonlinear scale correlator. Note that zero padding in (11) is not a compulsory, the experiments show that KST performs better than DSST in visual tracking even without the zero padding.

## Theoretical Analysis

The theoretical connection to the existing works, such as MOSSE (Bolme et al. 2010), KCF (Henriques, Caseiro, and Martins 2015), MKCF (Tang and Feng 2015), DSST (Danelljan et al. 2017) will be presented, respectively.

### Connection to MOSSE

The MOSSE filter also minimizes SSE and obtains the solution in Fourier domain. The difference is that MOSSE only finds a linear correlation filter, while the objective (2) of KCC is to find a non-linear correlator based on the kernel trick at a reasonable computational cost. More specifically, MOSSE is the linear case of KCC and KCC becomes MOSSE when $\boldsymbol{\kappa}_{\mathbf{z}}(\mathbf{z}) = \mathbf{z}$ in (1) and (4), so that the solution (12) to MOSSE is obtained,

$$\hat{\mathbf{h}}^* = \frac{\sum_{i=1}^{s} \hat{\mathbf{g}}^i \odot \hat{\mathbf{z}}^{i*}}{\sum_{i=1}^{s} \hat{\mathbf{z}}^i \odot \hat{\mathbf{z}}^{i*} + \lambda}. \quad (12)$$

Note that the solution (12) to MOSSE can only predict the translation of the signal $\mathbf{z}$ and the lengths of the filter and signals are always the same, i.e. $n = m$.

### Connection to KCF

KCF proposes a kernelized correlation filter but with many limitations. First, starting from the kernel regression model, the target output of KCF is a scalar, while KCC is based on the kernel cross-correlation, where target can be a vector or matrix. Second, as KCF is derived from the minimization of regression error in time domain, its solution in Fourier domain is obtained by diagonalizing the circulant matrix with discrete Fourier transform, which means that KCF can only be applied when the training samples are the circular shifts of one patch. On the contrary, the solution (4) to KCC is obtained by the minimization of correlation outputs error in Fourier domain naturally. This makes KCC applicable to arbitrary training samples. More specifically, KCF is a special case of KCC. To show this, assume only one training sample $\mathbf{z} \in \mathbb{R}^n$ is available, hence the solution (4) becomes (13).

$$\hat{\mathbf{h}}^* = \frac{\hat{\mathbf{g}}}{\hat{\boldsymbol{\kappa}}_{\mathbf{z}}(\mathbf{z}) + \lambda}. \quad (13)$$

If the sample-based vectors $\mathbf{z}_i$ are the circular shifts of the training sample $\mathbf{z}$, i.e. $\mathbf{z}_i \in \mathcal{T}_T(\mathbf{z})$, and $m = n$, the solution (13) is degraded to the solution to KTC with single training sample. It appears similar to the solution to KCF, where $\boldsymbol{\kappa}_{\mathbf{z}}(\mathbf{z})$ is replaced by the first row of a circulant kernel matrix and $\mathbf{g}$ is the stack vector of the ridge regression targets.

Another underlying problem of KCF is that it requires the kernel functions satisfy $\kappa(\mathbf{x}, \mathbf{z}) = \kappa(M\mathbf{x}, M\mathbf{z})$ for any permutation matrix $M$. However, KCC can be applied to any kernel function, which means that the properties of KCC are customizable with different kernel functions. This is useful for applications where some signal positions are more important than the others. For example, a kernel function can be designed to give more weights to those pixels near to the image center, so that the central part is focused. More properties can be designed according to the applications, however, it is out of the scope of this paper.

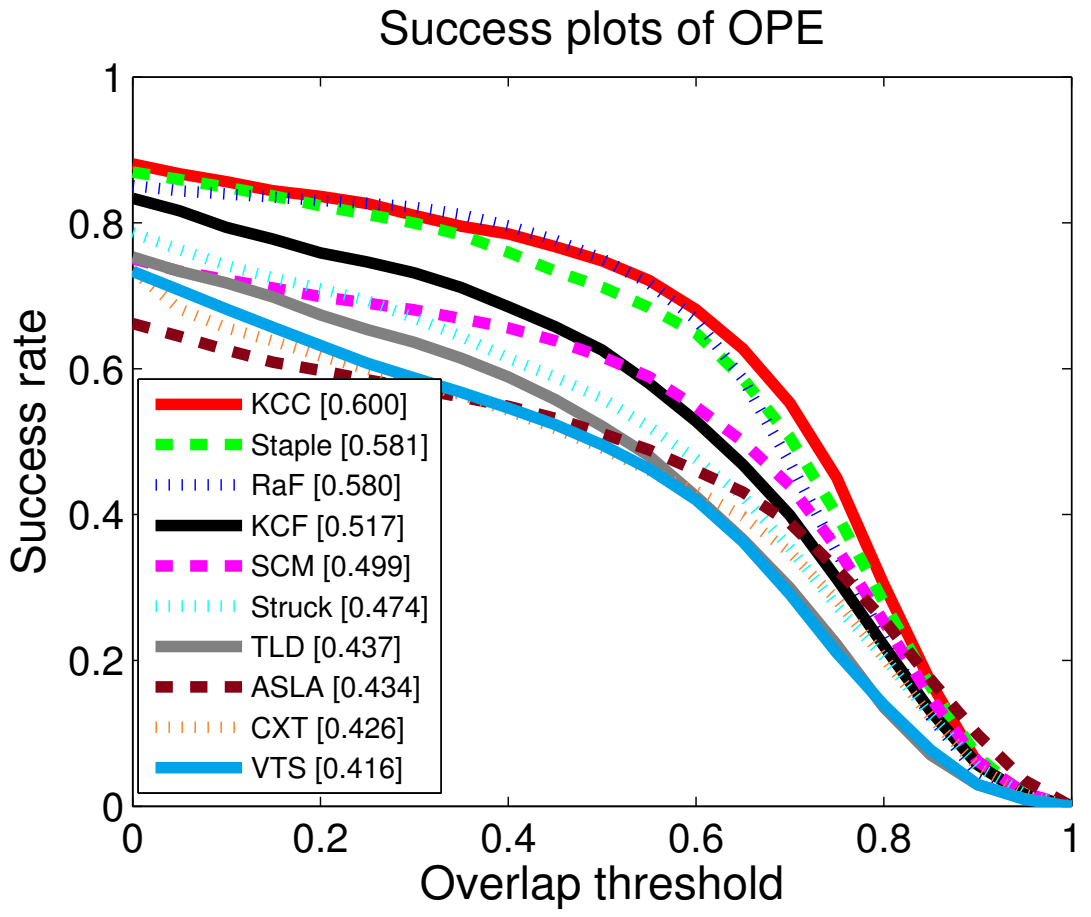

Figure 1: Success plot of different methods on all datasets.

In summary, KCF is a special case of KCC when the following conditions are satisfied: First, only one training sample is available; Second, the sample-based vectors are the circular translational shifts of the training sample; Third, only single kernel function is available; Last, the kernel function treats all dimensions of data equally.

### Connection to MKCF, DSST

MKCF extends KCF to multi-kernel case. It has been shown that KCC is degraded to the multi-kernel case when the $k_{th}$ element of the kernel vector is $\mathbf{d}^T\mathbf{k}_k$, where the vector $\mathbf{d}$ is the combination coefficients. Since MKCF is derived from KCF, all the limitations of KCF are imposed on MKCF. This limits its usages in other applications. Note that all the base kernels $\kappa_i$ for MKCF need to follow the circulant property mentioned in last section, which is not necessary for KCC.

DSST learns multiple MOSSE for scale tracking. In the detection step, several patches at different scales are sampled around the previous target. Then the scale transform of target is obtained by finding the best maximum response among all patches. In this sense, DSST extends MOSSE to multi-scale case at the price of repeatedly calculating the forward and backward FFTs at different scales. However, the proposed KSC can estimate the scale transform with single calculation of FFT, which reduces the complexity dramatically.

## Experiments

### Visual Tracking

Although the first CF is not proposed for visual tracking, it is visual tracking that ignites the boom of CFs. In this section, the performance of KCC on visual tracking will be shown.

**Implementation Details** In order to show the superiority of the proposed method over existing ones, we choose *Staple* (Sum of Template And Pixel-wise LEarners) (Bertinetto et al. 2016). It builds two CF based sub-trackers which can be used in conjunction to generate an inherently more robust tracker. In order to estimate the scale variations, Staple follows DSST and learns a distinct, multi-scale template for scale search using a 1-D correlation filter, at the price of repeatedly calculating the forward and backward FFTs at different scales. Here we show that this is not necessary and can be naturally solved by KSC. In both correlators, the regularization term $\lambda$ in (2) is set as $0.001$ and the Gaussian kernel is used with sigma of $0.1$.

**Dataset and Evaluation Metric** The well-known benchmark dataset that contains 51 sequences (Wu, Lim, and Yang 2013) is selected. It has been widely used for demonstration and comparison purpose. More importantly, it covers a variety of challenging scenarios. To better evaluate and analyze the strength and weakness of tracking approaches, the videos are categorized with 11 attributes based on different challenging factors including low resolution, in-plane and out-of-plane rotation, scale variation, occlusion, deformation, background clutters, illumination variation, motion blur, fast motion, and out-of-view. The standard criterion of success plot curve is adopted as the performance criteria. To measure the performance, the number of successful tracking frames whose overlap $S$ in (14) is larger than the given threshold $t_o$ is counted.

$$S = \left| \frac{r_t \bigcap r_g}{r_t \bigcup r_g} \right|, \tag{14}$$

where $\bigcap$ and $\bigcup$ stand for the intersection and union of two regions, respectively. $r_t$ and $r_g$ denote the bounding box of ground truth and tracking results, respectively. The success plot shows the ratios of successful frames at the thresholds varied from 0 to 1. The standard protocol to use the area under curve (AUC) of each success plot is followed to rank the tracking algorithms.

**Comparison** In order to demonstrate the performance of KCC, it is compared with all state-of-the-art methods in (Wu, Lim, and Yang 2013), KCF, and several recent advanced methods including Random Forest (Zhang et al. 2017) as well as Staple. In order to have a fair comparison, we only evaluate tracker with handcrafted features. Applying the proposed method on learning representation such as ConvNet features, however, is beyond the scope of this paper. Fig. 1 shows the success plot of different methods on all the 51 datasets under one pass evaluation (OPE) (Wu, Lim, and Yang 2013). In this case, a tracker is evaluated on a test sequence with initialization from the ground truth position in the first frame. As aforementioned, AUC is used to rank each method and for simplicity, only top 10 methods are illustrated. KCC performs significantly better than all the methods reported in (Wu, Lim, and Yang 2013), with an average relative improvement of $20\%$ with respect to the best tracker evaluated in the original benchmark (SCM (Zhong, Lu, and Yang 2014)). Moreover, KCC also outperforms recent trackers such as Random Forest (Zhang et al. 2017) and well studied diagnose framework (Wang et al. 2015b). Most importantly, KCC achieves better performance than Staple while running at roughly the same frame rate (40 FPS), which clearly demonstrates the superiority.

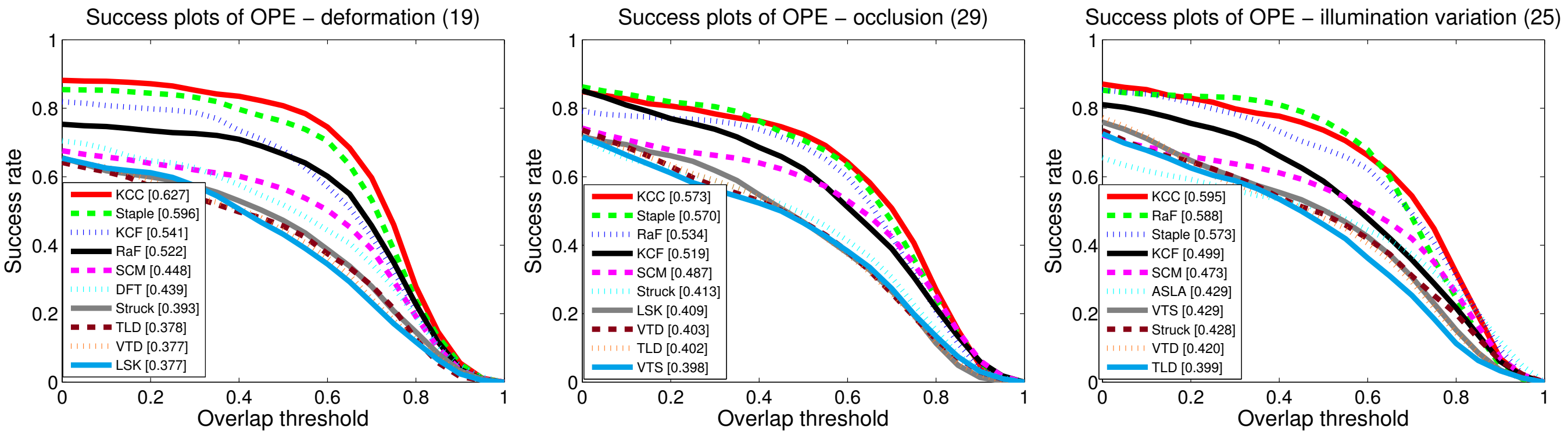

Figure 2: Success plot of different methods on challenging scenarios.

KCC performs well under a variety of challenging scenarios such as deformation, illumination variations, occlusions and so on. Due to page limits, only several representative examples are shown in Fig. 2, where the number in the caption indicates how many sequences belong to this scenario.

## Human Activity Recognition

Although KCC has shown its robustness and efficiency on visual tracking, its performance on non-visual signals is still unclear. To show the flexibility of KCC, its capability on human activity recognition using wearable devices is demonstrated. Interest in such devices has grown exponentially in recent years (Lara and Labrador 2013; Mosenia et al. 2017). Despite this popularity, barriers to adoption still remain.

**Dataset** The public wearable action recognition database (WARD) (Yang et al. 2009) is chosen. It is comprised of 20 human subjects (7 females and 13 males) with ages from 19 to 75 and 13 daily action categories, including rest at standing (ReSt), rest at sitting (ReSi), rest at lying (ReLi), walk forward (WaFo), walk forward left-circle (WaLe), walk forward right-circle (WaRi), turn left (TuLe), turn right (TuRi), go upstairs (Up), go downstairs (Down), jog (Jog), jump (Jump), and push wheelchair (Push). Five sensors, each of which consists of a triaxial accelerometer and a biaxial gyroscope, are located at the waist, left and right forearms, left and right ankles. Therefore, each sensor produces 5 dimensional data stream and totally 25 dimensional data is available. Each subject performs five trails for each activity, thus the database totally contains $20 \times 13 \times 5$ data stream, each of which lasts more than 10 seconds.

**Implementation Details** Since we are not aiming at establishing a comprehensive activity recognition system, but rather to demonstrate the robustness and flexibility of KCC, all the necessary data pre-processing procedures (Lara and Labrador 2013; Mosenia et al. 2017), such as noise removal, feature extraction, activity start time estimation, and PCA are skipped. Instead, the raw data stream are applied directly to KCC. Fig. 3 shows some of the data examples. We propose the KCC similarity $\mathcal{S}(\mathbf{x}, \mathbf{y})$ which is defined as the maximum value of KCC output in (15), that takes the signals $\mathbf{x}$ and $\mathbf{y}$ as the test and training samples, respectively.

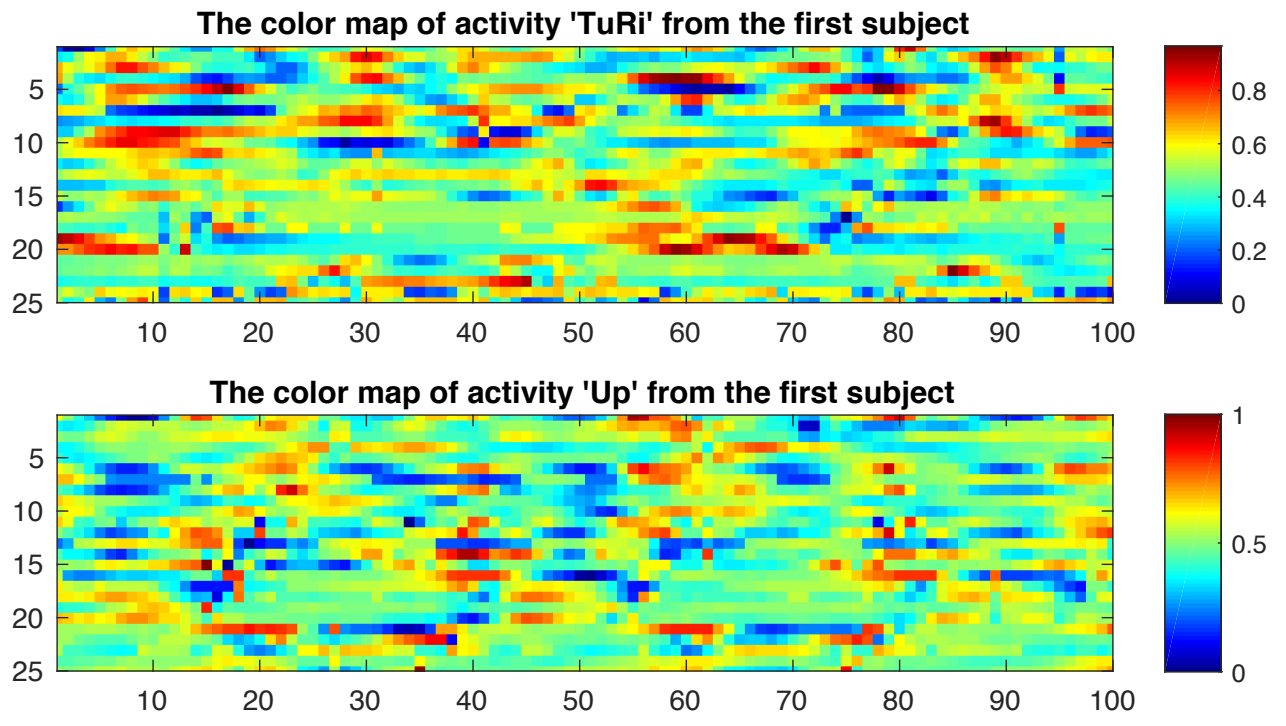

Figure 3: The color map examples of the 25 dimensional raw data from the wearable sensors. The horizontal axis is the time instant which is recorded with update rate 30Hz. The data are normalized for better visualization.

$$\boxed{\mathcal{S}(\mathbf{x}, \mathbf{y}) = \max \mathcal{F}^{-1}\left(\frac{\hat{\boldsymbol{\kappa}}_{\mathbf{y}}(\mathbf{x}) \odot \hat{\mathbf{g}}}{\hat{\boldsymbol{\kappa}}_{\mathbf{y}}(\mathbf{y}) + \lambda}\right)} \tag{15}$$

It is intuitive that the higher KCC simiarity is, the more similar two signals are. Therefore, KCC similarity can be applied for activity recognition based on the highest similarity value. In the experiments, short duration segments (50 instants, about $1.67$ s) are randomly selected from the database for each trail. Then the cross-validation with **single training sample** (one segment) is performed among all 13 activities. This process is repeated until each sample is taken as the training sample exactly one time, resulting a $65 \times 65$ confusion matrix, as there are in total $13 \times 5$ activity instances. Note that this process is more difficult than the 5-fold cross-validation, which takes 4 subsets in 5 as the training data. In the experiments, the regularization term $\lambda$ in (2) is set as $0.0015$ while the Gaussian kernel is used with sigma of $1$.

**Performance** The performance of KCC is compared with dynamic time warping (DTW). It is one of the most famous algorithms for measuring similarity between two temporal sequences which may vary in speed (Silva and Batista 2016). DTW might be one of the most suitable methods for comparison, since it has been widely used in sequence matching, e.g. speech recognition and data mining, etc (Senin 2008). Similar to the experiments for KCC, the classification is based on the shortest DTW distances among all the activities. Fig. 4 shows the confusion matrix for the two methods. It can be found that KCC performs much better in distinguishing different activities, and produces higher similarity measurements within classes. In summary, KCC achieves 90.7% overall classification accuracy with average running time 116.4µs, while DTW achieves 78.8% accuracy with average running time 156.9µs. This means than KCC improves 14.9% accuracy, while still saves 25.8% computational time. More details about the comparison for all 13 activities are shown in Fig. 5. It is noticed that both KCC and DTW cannot perform well on the first three activities, i.e. rest at standing, sitting, and lying. The possible reason is that the signals of the three activities are relatively flat and similar due to the static posture.

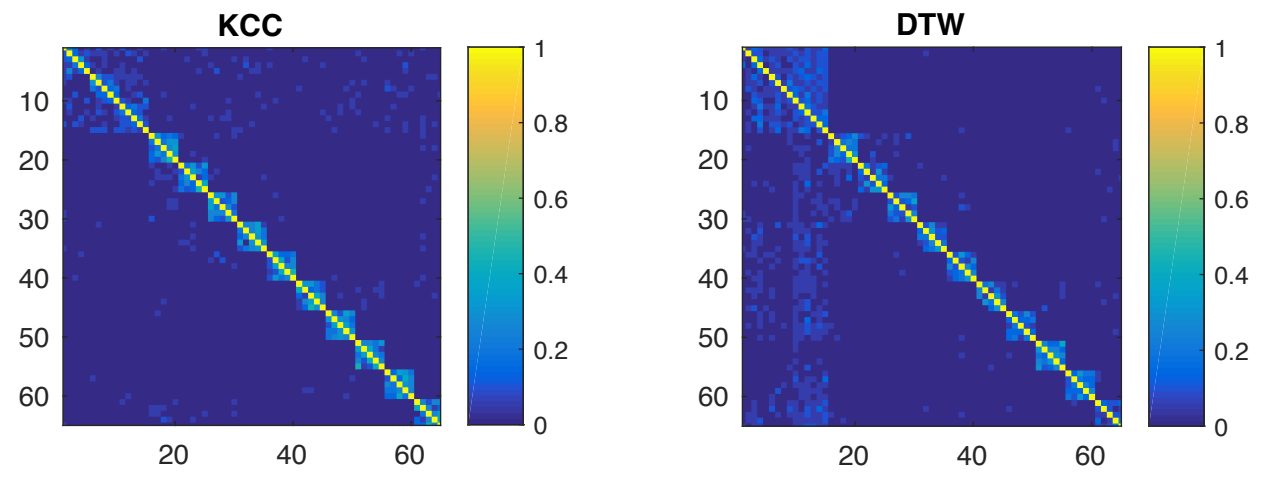

Figure 4: The average confusion matrix of KCC and DTW over all the subjects. There are 5 trials for each of the 13 activities, resulting in a $65 \times 65$ confusion matrix. It is obvious that KCC is able to better distinguish human activities and produce less false positives.

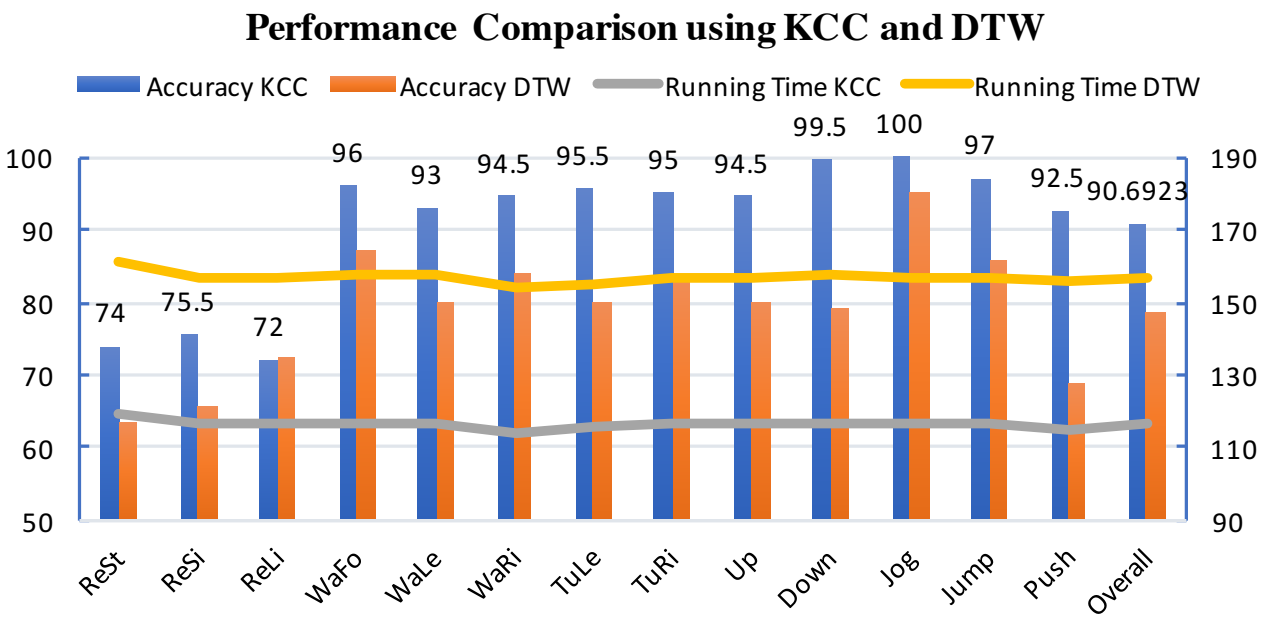

Figure 5: The performance comparison on WARD dataset. The left and right vertical axes indicate the recognition accuracy (%) and average running time (µs), respectively.

## Conclusion

A general formulation for non-linear correlator is proposed, which unifies the existing solutions. The advantages of KCC are its ability to be customized by specific kernels and not limited to circulant data structure, hence it can predict variations of a signal affine transformation, especially on scale, rotation and translation. The theoretical analysis of the KCC is conducted to demonstrate its generality and it is compared with the existing solutions to illustrate its flexibility. Besides showing the improvements in visual tracking, we extend its usages to another application, i.e. human activity recognition using wearable devices. Extensive experimental results on public datasets demonstrate that the proposed KCC performs favorably against the state-of-the-art methods in terms of efficiency, accuracy and robustness.

## Acknowledgments

The authors would like to thank Mr. Hoang Minh Chung, Xu Fang, and Junjun Wang for their help in the experiments.

## References

Avidan, S. 2007. Ensemble tracking. *IEEE Transactions on Pattern Analysis and Machine Intelligence* 29(2):261–271.

Babenko, B.; Yang, M.-H.; and Belongie, S. 2011. Robust object tracking with online multiple instance learning. *IEEE Transactions on Pattern Analysis and Machine Intelligence* 33(8):1619–1632.

Bertinetto, L.; Valmadre, J.; Golodetz, S.; Miksik, O.; and Torr, P. H. 2016. Staple: Complementary learners for real-time tracking. In *Proceedings of the IEEE Conference on Computer Vision and Pattern Recognition*, 1401–1409.

Bolme, D.; Beveridge, J. R.; Draper, B. A.; and Lui, Y. M. 2010. Visual Object Tracking Using Adaptive Correlation Filters. In *IEEE Conference on Computer Vision and Pattern Recognition*, 2544–2550. IEEE.

Bolme, D. S.; Draper, B. A.; and Beveridge, J. R. 2009. Average of Synthetic Exact Filters. In *IEEE Computer Society Conference on Computer Vision and Pattern Recognition Workshops*, 2105–2112. IEEE.

Casasent, D. 1984. Unified synthetic discriminant function computational formulation. *Applied Optics* 23(1):1620–1627.

Danelljan, M.; Khan, F. S.; Hager, G.; and Felsberg, M. 2017. Discriminative Scale Space Tracking. *IEEE Transactions on Pattern Analysis and Machine Intelligence* 1–1.

Fernandez, J. A.; Boddeti, V. N.; Rodriguez, A.; and Kumar, B. V. K. V. 2015. Zero-Aliasing Correlation Filters for Object Recognition. *IEEE Transactions on Pattern Analysis and Machine Intelligence* 37(8):1702–1715.

Galoogahi, H. K.; Sim, T.; and Lucey, S. 2015. Correlation Filters With Limited Boundaries. In *IEEE Conference on Computer Vision and Pattern Recognition*, 4630–4638.

Grabner, H.; Grabner, M.; and Bischof, H. 2006. Real-time tracking via on-line boosting. In *Proceedings of British Machine Vision Conference*, volume 1, 6.

Hare, S.; Saffari, A.; and Torr, P. H. 2011. Struck: Structured output tracking with kernels. In *2011 IEEE International Conference on Computer Vision*, 263–270. IEEE.

Henriques, J. F.; Caseiro, R.; and Martins, P. 2015. High-Speed Tracking with Kernelized Correlation Filters. *IEEE*

*Transactions on Pattern Analysis and Machine Intelligence* 37(3):583–596.

Hester, C. F., and Casasent, D. 1980. Multivariant technique for multiclass pattern recognition. *Applied Optics* 19(1):1758–1761.

Hong, Z.; Chen, Z.; Wang, C.; Mei, X.; Prokhorov, D.; and Tao, D. 2015. Multi-Store Tracker (MUSTer): a cognitive psychology inspired approach to object tracking. In *Proceedings of the IEEE Conference on Computer Vision and Pattern Recognition*, 749–758.

Kumar, B. V. K. V. 1986. Minimum-variance synthetic discriminant functions. *Journal of the Optical Society of America A* 3(10):1579–1584.

Lara, O. D., and Labrador, M. A. 2013. A Survey on Human Activity Recognition using Wearable Sensors. *IEEE Communications Surveys & Tutorials* 15(3):1192–1209.

Li, C.; Lin, L.; Zuo, W.; and Tang, J. 2017. Learning patch-based dynamic graph for visual tracking. In *AAAI*, 4126–4132.

Li, Y.; Zhu, J.; and Hoi, S. C. 2015. Reliable patch trackers: Robust visual tracking by exploiting reliable patches. In *Proceedings of the IEEE Conference on Computer Vision and Pattern Recognition*, 353–361.

Liu, T.; Wang, G.; and Yang, Q. 2015. Real-time part-based visual tracking via adaptive correlation filters. In *Proceeding of the IEEE Conference on Computer Vision and Pattern Recognition*, 4902–4912.

Ma, C.; Huang, J.-B.; Yang, X.; and Yang, M.-H. 2015a. Hierarchical convolutional features for visual tracking. In *Proceedings of the IEEE International Conference on Computer Vision*.

Ma, C.; Yang, X.; Zhang, C.; and Yang, M.-H. 2015b. Long-Term Correlation Tracking. In *IEEE Conference on Computer Vision and Pattern Recognition*, 5388–5396. IEEE.

Mahalanobis, A.; Vijaya Kumar, B. V. K.; Song, S.; Sims, S. R. F.; and Epperson, J. F. 1994. Unconstrained correlation filters. *Applied Optics* 33(1):3751–3759.

Mahalanobis, A.; Kumar, B. V. K. V.; and Casasent, D. P. 1987. Minimum average correlation energy filters. *Applied Optics* 26(1):3633–3640.

Mahalanobis, A.; Vijaya Kumar, B. V. K.; and Sims, S. R. F. 1996. Distance-classifier correlation filters for multiclass target recognition. *Applied Optics IP* 35(1):3127–.

Mei, X.; Ling, H.; Wu, Y.; Blasch, E. P.; and Bai, L. 2013. Efficient minimum error bounded particle resampling l1 tracker with occlusion detection. *IEEE Transactions on Image Processing* 22(7):2661–2675.

Mosenia, A.; Sur-Kolay, S.; Raghunathan, A.; and Jha, N. 2017. Wearable medical sensor-based system design: A survey. *IEEE Transactions on Multi-Scale Computing Systems*.

Refregier, P. 1991. Optimal trade-off filters for noise robustness, sharpness of the correlation peak, and Horner efficiency. *Optics Letters* 16(1):829–832.

Senin, P. 2008. Dynamic time warping algorithm review. *Information and Computer Science Department University of Hawaii at Manoa Honolulu, USA* 855:1–23.

Silva, D. F., and Batista, G. E. 2016. Speeding up all-pairwise dynamic time warping matrix calculation. In *Proceedings of the 2016 SIAM International Conference on Data Mining*, 837–845. SIAM.

Tang, M., and Feng, J. 2015. Multi-kernel Correlation Filter for Visual Tracking. In *IEEE International Conference on Computer Vision*, 3038–3046. IEEE.

Wang, N., and Yeung, D.-Y. 2013. Learning a deep compact image representation for visual tracking. In *Advances in Neural Information Processing Systems*, 809–817.

Wang, L.; Ouyang, W.; Wang, X.; and Lu, H. 2015a. Visual tracking with fully convolutional networks. In *IEEE International Conference on Computer Vision*, 3119–3127.

Wang, N.; Shi, J.; Yeung, D.-Y.; and Jia, J. 2015b. Understanding and diagnosing visual tracking systems. In *Proceedings of the IEEE International Conference on Computer Vision*, 3101–3109.

Wang, L.; Ouyang, W.; Wang, X.; and Lu, H. 2016. Stct: Sequentially training convolutional networks for visual tracking. In *Proceedings of the IEEE International Conference on Computer Vision*.

Wu, Y., and Huang, T. S. 2000. Self-supervised learning for visual tracking and recognition of human hand. In *AAAI/IAAI*, 243–248.

Wu, Y.; Lim, J.; and Yang, M.-H. 2013. Online object tracking: A benchmark. In *Proceedings of the IEEE Conference on Computer vision and pattern recognition*, 2411–2418. IEEE.

Yang, A. Y.; Jafari, R.; Sastry, S. S.; and Bajcsy, R. 2009. Distributed recognition of human actions using wearable motion sensor networks. *Journal of Ambient Intelligence and Smart Environments* 1(2):103–115.

Zhang, L., and Suganthan, P. N. 2016. Visual tracking with convolutional random vector functional link neural network. *IEEE Transactions on Cybernetics*.

Zhang, L., and Suganthan, P. N. 2017. Robust visual tracking via co-trained kernelized correlation filters. *Pattern Recognition* 69:82–93.

Zhang, K.; Liu, Q.; Wu, Y.; and Yang, M. H. 2016. Robust visual tracking via convolutional networks without training. *IEEE Transactions on Image Processing* 25(4):1779–1792.

Zhang, L.; Varadarajan, J.; Suganthan, P. N.; Ahuja, N.; and Moulin, P. 2017. Robust visual tracking using oblique random forests. In *IEEE International Conference on Computer Vision and Pattern Recognition*. IEEE.

Zhong, W.; Lu, H.; and Yang, M.-H. 2014. Robust object tracking via sparse collaborative appearance model. *IEEE Transactions on Image Processing* 23(5):2356–2368.